# Enhancing the Product Quality of the Injection Process Using eXplainable Artificial Intelligence

Jisoo Hong [1], Yongmin Hong [1], Jung-Woo Baek [2] and Sung-Woo Kang [1,*]

1. Department of Industrial Engineering, Inha University, Incheon 22212, Republic of Korea; wltnghd5182@inha.edu(J.H.); hym9771@gmail.com(Y.H);
2. Department of Industrial & Systems Engineering, Dongguk University, Seoul 04620, Republic of Korea; jwbaek@dongguk.edu
* Correspondence: kangsungwoo@inha.ac.kr

**Abstract:** The injection molding process is a traditional technique for making products in various industries such as electronics and automobiles via solidifying liquid resin into certain molds. Although the process is not related to creating the main part of engines or semiconductors, this manufacturing methodology sets the final form of the products. Recently, research has continued to reduce the defect rate of the injection molding process. This study proposes an optimal injection molding process control system to reduce the defect rate of injection molding products with XAI (eXplainable Artificial Intelligence) approaches. Boosting algorithms (XGBoost and LightGBM) are used as tree-based classifiers for predicting whether each product is normal or defective. The main features to control the process for improving the product are extracted by SHapley Additive exPlanations, while the individual conditional expectation analyzes the optimal control range of these extracted features. To validate the methodology presented in this work, the actual injection molding AI manufacturing dataset provided by KAMP (Korea AI Manufacturing Platform) is employed for the case study. The results reveal that the defect rate decreases from 1.00% (Original defect rate) to 0.21% with XGBoost and 0.13% with LightGBM, respectively.

## 1. Introduction

During the injection molding process, liquid raw materials are injected into a mold and hardened to produce a product. It is widely used as an effective technique to mass-produce large core components and small parts, such as automobiles, displays, and semiconductors. The injection molding process maintains a relatively high quality and has been improved over time.

Injection molding manufacturers have recently employed machine learning, deep learning, and artificial intelligence to the injection molding process [1-4]. However, machine learning and deep learning often lack transparency and interpretability, making them unfamiliar to field operators.

The injection molding process has been continuously improved hereby reaching a high yield rate.(over 90%) However, achieving a process yield close to 100% from an already high-yield state requires fine-tuning of process variables. This paper aims to reduce the defect rate of injection-molded products, by employing eXplainable Artificial Intelligence (XAI) algorithm to fine-tune the process variables.

Traditional machine learning techniques that exhibit black-box characteristics, lack the ability to provide explanations for their predictions, thereby demonstrating limited reliability. This shortcoming poses significant challenges to their practical implementation in real-world processes. However, XAI methods provide clear reasons and justifications for the model's outcomes. This feature makes XAI a suitable approach for fine-tuning process variables to improve the defect rate in injection molding processes. This paper aims to enhance the reliability of the process and achieve even higher yield rates by employing XAI. Also, XAI enables field experts to more easily understand AI predictions by providing evidence for model learning.

SHAP (SHapley Additive exPlanations) extracts the main features affecting product defects. Tree-based algorithms, such as XGBoost and LightGBM, are used as training models for feature extraction. The optimal control range of features identified through SHAP is determined using the ICE (Individual Conditional Expectation) algorithm.

The remainder of this paper is organized as follows. Section 1 introduces the motivation and purpose of this study. Section 2 describes previous studies. Section 3 presents a methodology that explains the process management method presented in this paper. Section 4 presents the experimental results using actual injection molding process data. Section 5 discusses the conclusions and future work.

## 2. Related Studies

*2.1. Injection Process*

The injection molding process involves plastic molding. The structure of injection molding process is shown in Figure 1.

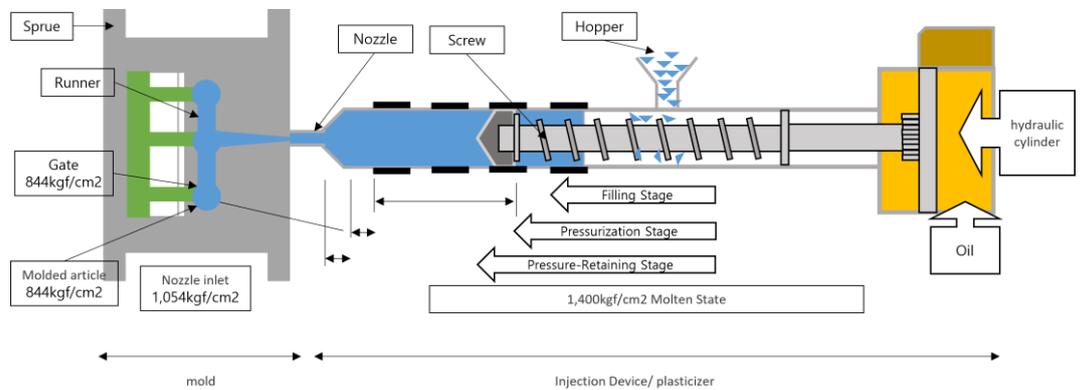

**Figure 1.** Structure of Injection Molding Process

The injection process involves plastic molding. This process is performed by injecting a dissolved thermoplastic resin into a mold and cooling it[5].

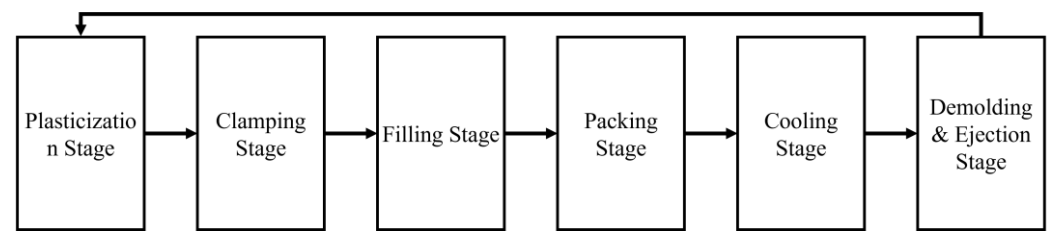

**Figure 2.** Injection Process

The injection molding process has six stages, as shown in Figure 2: plasticization, clamping, filling, packing, cooling, demolding, and ejection [6].

1. Plasticization stage: The screw moves forward, and the plastic resin is dissolved by a heated barrel.

2. Clamping stage: The oil pressure system enables the plastic resin to fit the fixed and movable parts of the mold closely.

3. Filling stage: The mold is filled with dissolved plastic resin from the nozzle.

4. Packing stage: To prevent the volume from shrinking, pressure was applied before the plastic resin hardens completely.

5. Cooling stage: The dissolved plastic resin is cooled and hardened.

6. Demolding and ejection stage: When the mold is opened, the resin shrinks, and the product is ejected.

The injection molding products are processed by repeating the clamping, demolding, and ejection stages. Because the injection molding process produces finished products, a high quality must be maintained. Therefore, the optimal management of variables, such as temperature and pressure, which are the major variables that determine product quality, is very important for improving the process product yield.

Controlling the parameters of the injection molding process is important for optimization in various fields. In the field of injection molding process control for internal combustion engines, numerical analysis of the injection molding process is performed by modeling and computer simulations based on multiple fuel injections[7]. The AVL Boost simulation application is used to monitor engine functionality. However, the simulation used only three monitoring conditions. This study uses continuous feature conditions to propose the control range of main features. In the medical field, research on injection molding process optimization is also being conducted. A polycaprolactone parts development system is proposed for future implants through several injection molding parameter improvements, including the melting temperature, injection time, and injection pressure[8]. The results of this system demonstrate the potential of using simulations as tools to optimize the injection-molding process. However, the data used in this study are artificial data generated from the literature. Therefore, it is necessary to consider its application in actual processes.

Injection molding process has low defect rate. Therefore, failure data is extremely lower than the normal product data. Consequently, when applying artificial intelligence to injection molding process data, an imbalance between normal and defective data is inherent. Various studies have been conducted to address this issue [9-11]. SMOTE(Synthetic Minority Over-sampling TechniquE) is appropriate for addressing data imbalance in manufacturing processes because it generates new data points between existing variable values[9]. This study employs the SMOTE technique to augment defective data, thereby resolving the imbalance problem.

*2.2. eXplainable Artificial Intelligence(XAI)*

Unlike existing AI, explainable XAI is a algorithm that increases reliability by presenting validity and grounds for machine learning[12]. Original AI has the "black box" characteristic that does not provide grounds for prediction results. In 2017, the Defense Advanced Research Projects Agency suggested using XAI to address the limitations of AI, as shown in Figure 3 [13]. Because of these characteristics of XAI, field experts can easily understand the prediction results.

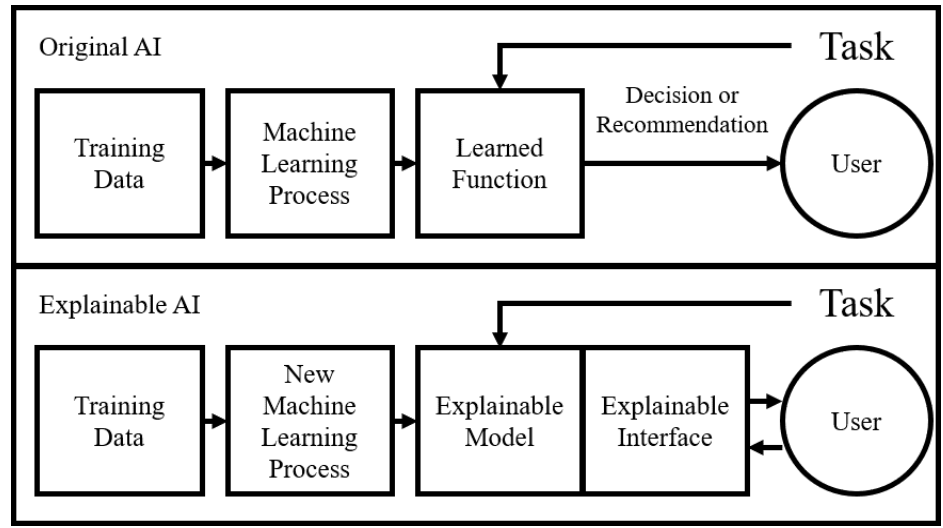

**Figure 3.** eXplainable Artificial Intelligence(XAI)

Recently, research into yield improvement processes based on these factors has progressed. Zhang proposed a fault-diagnosis system for oil-immersed transformers [14]. The system used the SHAP for feature selection and achieved a recall value of 0.96 for the fault samples[15]. However, no additional measures were conducted for the selected features. This study employs ICE algorithm to provide the optimal control range of each selected features to the field experts.

To improve manufacturing quality, rule-based explanations are performed based on ensemble machine learning[16]. Feature importance is used to obtain the most significant process conditions, and PDP(Partial Dependence Plot) and ICE plots are used to provide a visual overview. However, the feature importance does not consider the correlation of each feature. The SHAP algorithm creates a subset of each feature to extract the main features by calculating the correlations. In addition, this study uses the PDP and ICE plots to determine the optimal control range of the main features.

## 3. Methodology

The injection molding process is a traditional manufacturing method with high production yield. This process is the final step in creating the surface of a product. Therefore, it is directly related to product defects, and strict yield management is required. Recently, XAI has become a state-of-the-art methodology for improving manufacturing processes. This paper presents a pilot study for implementing XAI to increase the injection molding process yield. This study aims to improve the injection molding process based on artificial intelligence, and the methodology of the study is shown in Figure 4.

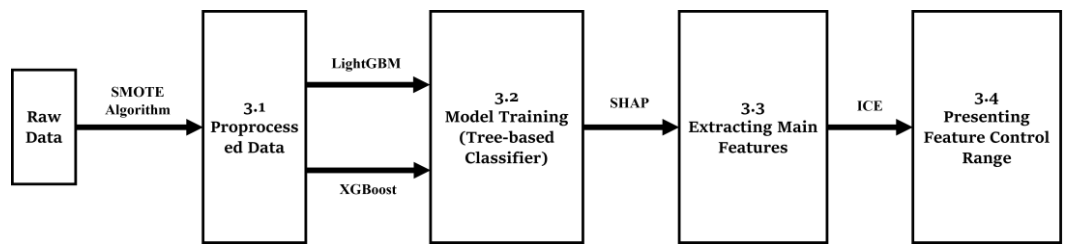

**Figure 4.** Flowchart of the Methodology

The injection process shows a data imbalance between normal and defect data owing to the high yield of its own nature. To resolve the data imbalance, the SMOTE technique is employed in the data preprocessing stage. (Section 3.1) Then, the tree-based classifier (Section 3.2) trains a model for predicting the product's defect. The SHAP Algorithm (Section 3.3) extracts major features that critically affect defect prediction. Finally, the control range of the major features is determined using the ICE algorithm (Section 3.4).

*3.1. Data Preprocessing for Injection Process*

This study uses the injection molding process data collected by sensors from a mold and machine[17]. The DataFrame is constructed by selecting controllable features such as temperature and pressure. The injection molding process has a high yield; therefore, the numbers of normal data and defect data are imbalanced, which results in a biased analysis. Therefore, oversampling is performed to balance the data used in the study. To solve this problem, this study employs the SMOTE algorithm for oversampling. SMOTE is a k-nearest neighbor (KNN)-based oversampling algorithm[18]. Figure 5 shows the operating principle of SMOTE.

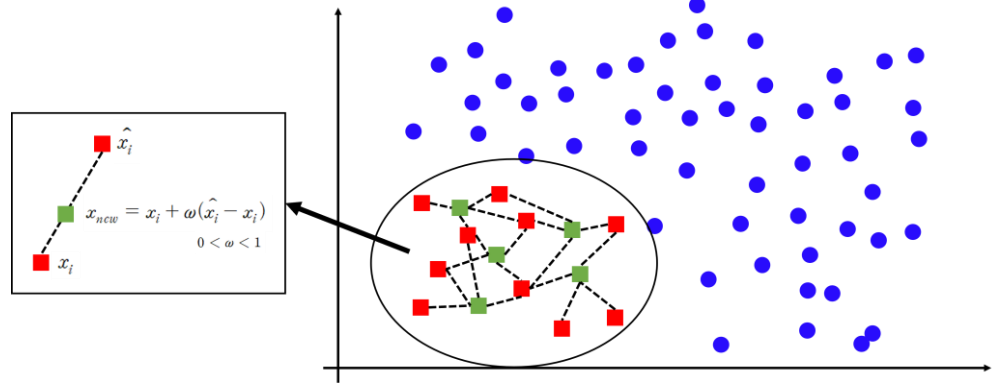

**Figure 5.** Operating Principle of the SMOTE Algorithm

First, one selects one of the data points of the minority class; in this case, the defect is a minority class, such as the red squares ($x_i$) in Figure 5. The squares represent defect data for the injection molding process. One of the K nearest data points of the corresponding data is randomly selected, and the difference between the two selected data points is multiplied by the weight to generate new data, such as the green squares in Figure 5($x_{new}$). In this case, the weight is randomly generated between zero and one. The imbalance in the data is resolved by repeating this process until a sufficient amount of data is generated. In this study, defective data are oversampled to equal the amount of normal data. Because the injection molding process data is distributed within a similar range owing to the characteristics of the process, the SMOTE algorithm is employed to generate virtual defect datasets close to the original data.

*3.2. Tree Based Classifier(XGBoost, LightGBM)*

This study uses a tree-based classifier to learn and predict whether products are defective. The tree-based classifiers used in this study are XGBoost and LightGBM. XGBoost is a gradient-boosting-based algorithm that combines several weak decision trees to build a robust model[19,20]. XGBoost is widely used in many ways because of its parallel learning, fast calculation speed, and excellent performance. The learning process for XGBoost is shown in Table 1.

**Table 1.** XGBoost Algorithm

| XGBoost (eXtreme Gradient Boosting) |
| --- |

*Input:*

Instance set of current node; feature dimension;

*Procedure:*

$J(P) = 0$

$G = \sum_{i \in I} g_i, H = \sum_{i \in I} h_i$

$for\ k = 1\ to\ n\ do$

$\quad G_L = 0, H_L = 0$

$\quad for\ j\ in\ sorted\ do$

$\quad\quad G_L = G_L + g_j, H_L = H_L + H_j$

$\quad\quad G_R = G - G_L, H_R = H_L - H_L$

$\quad\quad score = \max(score, J(P))$

$\quad end$

$end$

*Output:* Split with max score

LightGBM is a gradient-boost-based algorithm, like XGBoost[21, 22]. The primary technology used is gradient-based one-sided sampling (GOSS), which applies multiplier constants to low-weight objects. LightGBM uses memory more efficiently by dividing the tree leafwise rather than levelwise; therefore, it exhibits good speed and performance. A levelwise tree requires additional operations to balance it. However, a leafwise tree is more efficient, because it divides and calculates the node with the largest delta loss. The LightGBM learning process is shown in Table 2.

**Table 2.** LightGBM Algorithm

**LightGBM (Light Gradient Boosting Machine)**

*Input:*

$Training\ data$:

$\boldsymbol{D} = \{(x_1, y_1), (x_2, y_2), \ldots, (x_N, y_N)\}$,

$x_i \in x, x \subseteq R, y_i \in -1, +1$;

$Loss\ function: L(y, \theta(x))$

*Iterations:*

$\boldsymbol{M}$; Big gradient data sampling ratio: a;

slight gradient data sampling ratio: b;

1. $Combine\ features\ that\ are\ mutually$

$exclusive (i.e., features\ never\ simultaneously$

$accept\ nonzero\ values)\ of\ x_i, i = \{1, \ldots, N\}\ by$

$the\ exclusice\ feature\ bundling\ (EFB)\ technique;$

2. $Set\ \theta_0(x) = argmin_c \sum_{i}^{N} L(y_i, c);$

3. $for\ m = 1\ to\ M\ do$

4. $Calculate\ gradient\ absolute\ values$;

$r_i = |\partial L(y_i, \theta(x_i))/\partial \theta(x_i)|_{\theta(x)=\theta_{m-1}(x)}, \quad i = \{1, \dots, N\}$

5. $Resample\ data\ set\ using\ gradient\ based\ one\ side\ sampling\ (GOSS)\ process$;

$topN = a \times len(D); randN = b \times len(D);$

$Sorted = GetSortedIndices(abs(r));$

$A = sorted[1:topN];$

$B = RandomPick(sorted[topN:len(D)], randN);$

$\acute{D} = A + B;$

6. $Calculate\ information\ gains$;

$$Vj(d) = \left(\left(\sum_{x_i \in A_l} r_i + ((1-a)/b) \sum_{x_i \in B_l} r_i\right)^2 / n_l^j(d)\right.$$

$$\left. + \left(\sum_{x_i \in A_r} r_i + ((1-a)/b) \sum_{x_i \in B_r} r_i\right)^2 / n_r^j(d)\right)/n$$

7. $Develop\ a\ new\ decision\ tree\ \theta_m(x)'\ on\ set\ D'$

8. $Update\ \theta_m(x) = \theta_{m-1}(x) + \theta_m(x)$

9. $End$

**Output:** Return $\tilde{\theta}(x) = \theta_M(x)$

*3.3. Shapley Additive exPlanations (SHAP)*

The SHAP algorithm extracts the main features of the injection molding process by exploring the impact of each feature on product quality. The algorithm is based on Shapley's game theory, which examines how individuals make decisions when faced with interdependent circumstances. This algorithm regards each manufacturing feature as an individual in game theory. The impact on feature i is analyzed using the process described in Figure 6.

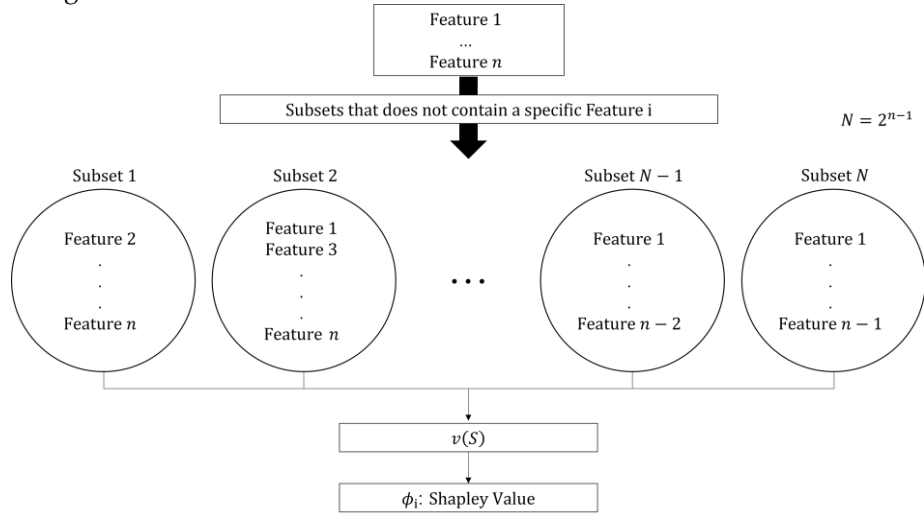

**Figure 6.** Procedure for Obtaining the Shapley Value

$$v(S) = \int \hat{f}(x_1, \ldots, x_n) dP_{x \notin S} - E_x(\hat{f}(X)) \quad (1)$$

$$\phi_i(v) = \sum_{S \subseteq 1,\ldots,n \{i\}} \frac{|S|!\,(n-|S|-1)!}{n!} (v(S \cup \{i\}) - v(S)) \quad (2)$$

$\phi_i$: Shapley Value for manufacturing feature i

n: Total number of manufacturing features

S: Subset that does not contain manufacturing feature i

v(S) : Contribution of a subset S

v(S∪i): Contribution of a subset (S∪i)

The SHAP algorithm generates every possible subset of each manufacturing feature. To examine the influence of a manufacturing feature, one subtracts the algorithm subsets the contribution of a subset which does not contain features from the contribution of a subset; the contribution of the subset is calculated as shown in (1). To check the importance of the feature, as shown in (2), a value called the Shapley value is calculated. In this study, the Shapley values are used to select the main features. The mean absolute Shapley Value is used to consider both the negative and positive influences on the product. Figure 7 shows the Shapley Value for each instance and expresses the mean of the absolute Shapley Value. The SHAP algorithm addresses the limitations of traditional variable importance methods (e.g., Feature Importance) by accounting for both negative and positive interactions between variables.

The injection features are sorted in descending order of importance. The main features of the process are selected based on the line in which the cumulative importance of the features is 70% of the total importance.

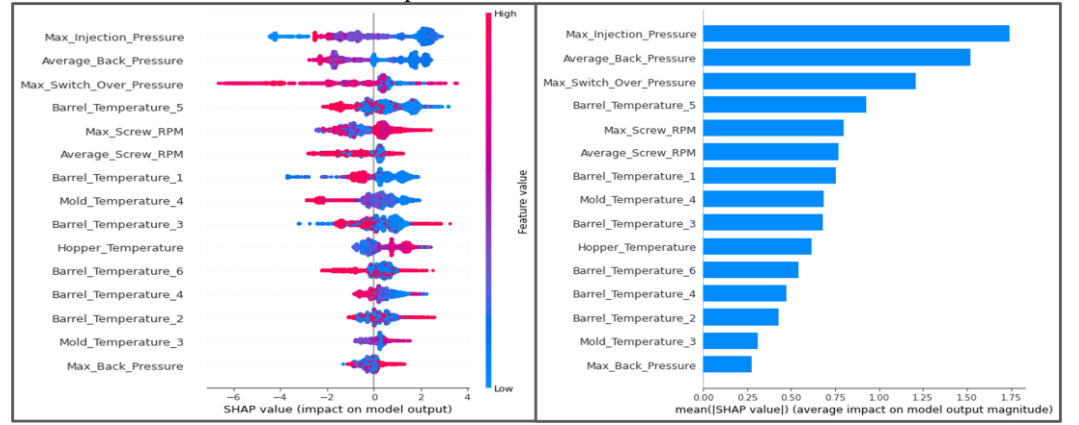

**Figure 7.** Representative Plots of the SHAP Value

*3.4. ICE and PDP*

To explore the conditions for improving the injection quality, both the ICE and PDP algorithms are proposed to determine the control range of the main features. The ICE predicts the target value of an instance according to the changes in the feature values of the manufacturing process. In the injection molding process, the target value is predicted by fixing other features (temperature and RPM) and changing a particular feature (pressure) to propose a control pressure range. The ICE process is presented in Table 3.

**Table 3.** Procedure Used by the ICE Algorithm to Predict the Control Range in the Injection Process

| ICE algorithm to predict the control range in injection molding process |
|---|

**Input:**

$X_i$ : A specific manufacturing feature for presenting the control range

$X_i'$ : All manufacturing features except $X_i$

$N$ : Number of instance

$p, q$ : Each instance

**Procedure:**

1. Initialize model with a constant value

$\hat{f}(X_i^{(p)}, X_i^{(q)'})_{p,q=1}^N$

2. for $q = 1$ to $N$:

    for $p = 1$ to $N$:

        $X_i^p$ = The value of $X_i$ in index $p$

        $X_i^{q'}$ = The value of $X_i'$ in index $q$

Plotting $\hat{f}(X_i^{(p)}, X_i^{(q)'})$

**Output:** ICE & PDP plot

## 4. Experimental Results

This paper aims to present a process yield improvement methodology using XAI-based algorithms. The main features are derived using SHAP, and their control range is determined using ICE.

### 4.1. Collection and Preprocessing for the Injection Process

This study uses automobile windshield side molding injection molding process data collected from October 16th, 2020 to November 19th, 2020. The total number of collected data points is 7,990, and the number of features is 45. Total dataframe is shown in Table Ⅳ. The target value is "PassOrFail," and it is expressed as 1 for normal products and 0 for defective products.

**Table 4.** Example of Injection Process Dataset.

| PassOFail | Average_Screw_RPM | Max_Screw_RPM | Barrel_Temperature_1 | … | Max_Injection_Pressure |
|---|---|---|---|---|---|
| 1 | 292.5 | 30.7 | 276.5 | … | 141.8 |
| 1 | 292.4 | 30.8 | 276.2 | … | 141.7 |
| 1 | 292.5 | 30.8 | 276.2 | … | 141.7 |
| 1 | 292.6 | 31.0 | 276.5 | … | 141.5 |
| 1 | 292.6 | 30.8 | 276.8 | … | 142.5 |
| 0 | 292.5 | 30.9 | 276.3 | … | 142.6 |
| 1 | 292.5 | 31.0 | 275.5 | … | 142.5 |
| ⋮ | ⋮ | ⋮ | ⋮ | ⋮ | ⋮ |
| 0 | 290.5 | 30.9 | 286.1 | … | 142.6 |

The preprocessing is performed in three steps. A dataframe is constructed by selecting 16 controllable features such as temperature, pressure, and RPM from the collected process features. Time features such as 'Filling_Time', 'Ejection_Time' and position features are excluded due to uncontrollability. Also, products with different process indices are excluded as they violate the control variables. Subsequently, a process is conducted to check for missing values or outliers. An example of the selected process features is presented in Table 5.

**Table 5.** Independent Variables of the Injection Molding Process Data.

| Independent Variable (Unit) | Description |
| --- | --- |
| Max_Screw_RPM (mm/s) | Maximum speed of screw for injection |
| Average_Screw_RPM (mm/s) | Average speed of screw for injection |
| Max_Injection_Pressure (MPa) | Maximum pressure applied to the molten resin flowing into the mold |
| Max_Switch_Over_Pressure (MPa) | Pressure converted from injection to packing pressure |
| Average_Back_Pressure (MPa) | Average pressure to prevent the screw from being pushed out |
| Barrel_Temperature_1~7 (°C) | Temperature of the barrel |
| Hopper_Temperature (°C) | Temperature of the hopper |
| Mold_Temperature_3, 4 (°C) | Temperature of the mold |

Training and validation are performed using train–test splits. The training and test datasets are split in a 5:5 ratio, and each split dataset is listed in Table 6.

**Table 6.** Result of the Train-Test Split

|  | Normal | Defective |
| --- | --- | --- |
| Train Dataset | 3,964 | 31 |
| Test Dataset | 3,955 | 40 |

The SMOTE algorithm is used to balance the ratios of normal and defective data. The results of the oversampling are listed in Table 7.

**Table 7.** Oversampling Results

|  | Normal | Defective |
| --- | --- | --- |
| Train Dataset | 3,964 | 3,964 |
| Test Dataset | 3,955 | 40 |

*4.2. Model Training for Injection Process*

This study uses a tree-based classifier, XGBoost, and LightGBM to train and predict whether injection molding process products are defective. The training dataset (Normal Data: 3964 / Defective Data: 3964) is used for training, and the Test Dataset (Normal Data: 3955 / Defective Data: 40) is used to check the accuracy of the model. Additionally, cross-validation is performed to check the model's performance. During the cross-validation process, the number of subsets is set to three. For XGBoost, the accuracy of each cross-validation is 0.9947, 0.9977, and 0.9981, with a CV average accuracy of 0.9968. For LightGBM, the respective accuracies are 0.9924, 0.9955, and 0.9977, with a CV average accuracy of 0.9952. The results of XGBoost and LightGBM are presented in Table 8.

**Table 8.** Model Training Results

|  |  | Actual Normal Data | Actual Defective Data | Accuracy | CV Average Accuracy |
|---|---|---|---|---|---|
| XGBoost | Predicted Normal Data | 3,941 | 25 | 99.02 | 0.9968 |
|  | Predicted Defective Data | 14 | 15 |  |  |
| LightGBM | Predicted Normal Data | 3,941 | 25 | 99.02 | 0.9952 |
|  | Predicted Defective Data | 14 | 15 |  |  |

### 4.3. SHAP(Shapley Additive exPlanations)

To verify the importance of features in the injection molding process, the main features are extracted by using the SHAP algorithm. Figure 8 shows the mean absolute Shapley value of each manufacturing feature for XGBoost and LightGBM.

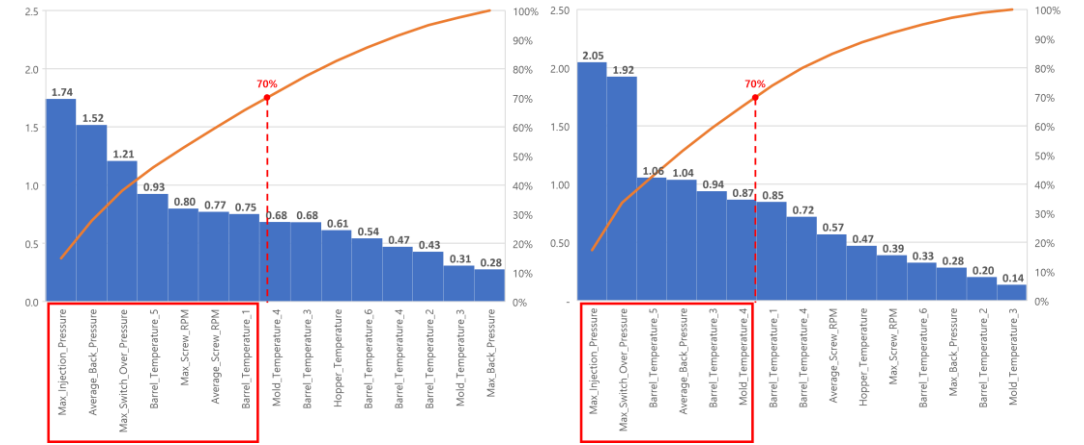

**Figure 8.** Shapley Value of Manufacturing Features (Left: XGBoost, Right: LightGBM)

Each graph shows the importance of manufacturing features in descending order. Features with cumulative importance corresponding to 70% of the total are selected as the main features. In the case of XGBoost, the main features are "Max Injection Pressure," "Average Back Pressure," "Max Switch Over Pressure," "Barrel Temperature 5," "Max Screw RPM," "Average Screw RPM," and "Barrel Temperature 1."

In the case of LightGBM, the main features are "Max Injection Pressure," "Max Switch Over Pressure," "Barrel Temperature 5," "Average Back Pressure," "Barrel Temperature 3," and "Mold Temperature 4." The selected main features and mean absolute Shapley values are listed in Table 9.

**Table 9.** Selected Main Features and Mean of the Absolute Shapley Value

|  | XGBoost | | Cumulative Ratio |
|---|---|---|---|
|  | Feature Name | Value |  |
| 1 | Max_Injection_Pressure | 1.74 | 0.15 |
| 2 | Average_Back_Pressure | 1.52 | 0.28 |
| 3 | Max_Switch_Over_Pressure | 1.21 | 0.38 |
| 4 | Barrel_Temperature_5 | 0.93 | 0.46 |
| 5 | Max_Screw_RPM | 0.80 | 0.53 |
| 6 | Average_Screw_RPM | 0.77 | 0.59 |
| 7 | Barrel_Temperature_1 | 0.75 | 0.66 |
| | LightGBM | | Cumulative Ratio |
| | Feature Name | Value | |
| 1 | Max_Injection_Pressure | 2.05 | 0.17 |
| 2 | Max_Switch_Over_Pressure | 1.92 | 0.34 |
| 3 | Barrel_Temperature_5 | 1.06 | 0.43 |
| 4 | Average_Back_Pressure | 1.04 | 0.51 |
| 5 | Barrel_Temperature_3 | 0.94 | 0.59 |
| 6 | Mold_Temperature_4 | 0.87 | 0.67 |

*4.4. ICE and PDP*

The ICE algorithm extracts the control range of the main features to reduce the process-defect rate. The ICE plots of the main features selected in Section 4.3 by each XGBoost and LightGBM, are given in Figure 9 and 10, respectively.

Each control range of the main features is presented according to the algorithm described in Section 3.4. The PDP is the average of the ICE experimental results, which are represented by orange dotted lines in Figures 9 and 10. The minimum and maximum PDP values of each main feature are indicated by red lines in Figures 9 and 10.

For example, in the case of Figure 10 (b), the maximum PDP value is 0.73, and the minimum value is 0.26. Both values are calculated according to the change in the x value Max_Switch_Over_Pressure. Tables 10 and 11 show the control ranges of the main features for alpha values of 0.05, 0.1, and 0.2 based on the y-axis maximum values.

**Table 10.** Control Range of the Main Features for Three Alpha Values (XGBoost Results)

| Variable $\alpha$ | 0.05 | 0.1 | 0.2 |
|---|---|---|---|
| Max_Injection_Pressure | [141.60, 142.40] | [141.20, 183.20] | [141.20, 183.20] |
| Average_Back_Pressure | [13.30, 90.80] | [13.30, 90.80] | [13.30, 90.80] |
| Max_Switch_Over_Pressure | [115.60, 136.50] | [115.60, 136.52] | [115.60, 136.52] |
| Barrel_Temperature_5 | [236.30, 255.00] | [236.30, 266.40] | [236.30, 266.40] |
| Max_Screw_RPM | [30.30, 31.20] | [30.30, 31.20] | [30.30, 31.20] |
| Average_Screw_RPM | [29.00, 293.40] | [29.00, 293.40] | [29.00, 293.40] |
| Barrel_Temperature_1 | [244.70, 287.10] | [244.70, 287.10] | [244.70, 287.10] |

**Table 11.** Control Range of the Main Features for Three Alpha Values (LightGBM Results)

| Variable $\alpha$ | 0.05 | 0.1 | 0.2 |
|---|---|---|---|
| Max_Injection_Pressure | [141.50, 142.20] | [141.20, 183.20] | [141.20, 183.20] |
| Max_Switch_Over_Pressure | [115.60, 119.00] | [115.60, 119.55] | [115.60, 136.80] |
| Barrel_Temperature_5 | [236.30, 254.90] | [236.30, 255.00] | [236.30, 266.40] |
| Average_Back_Pressure | [13.30, 60.00] | [13.30, 60.00] | [13.30, 60.00] |
| Barrel_Temperature_3 | [285.50, 285.80] | [245.00, 285.40] | [245.00, 285.40] |
| Barrel_Temperature_4 | [20.60, 22.60] | [20.60, 22.69] | [20.60, 27.70] |

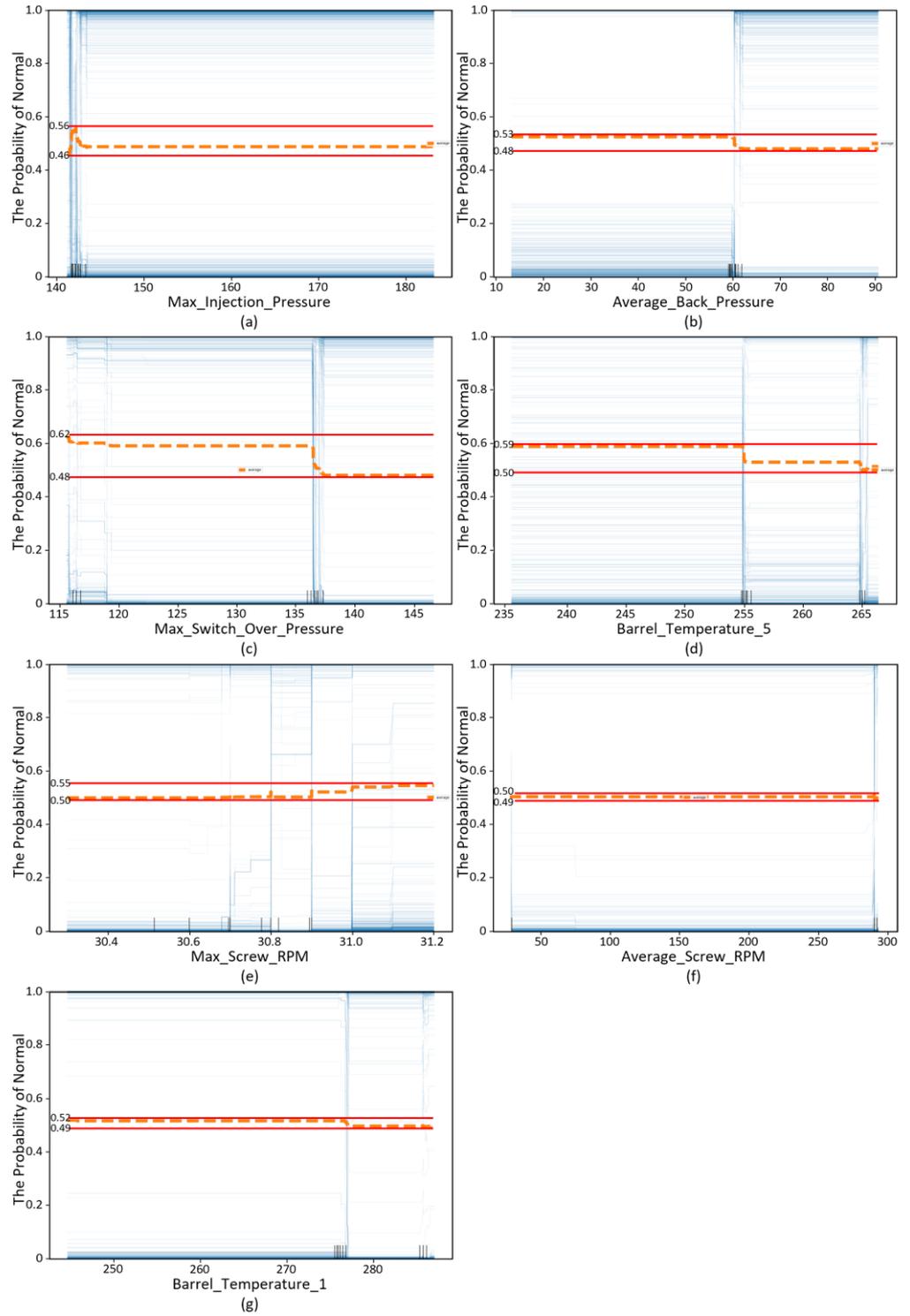

**Figure 9.** ICE Plots of XGBoost

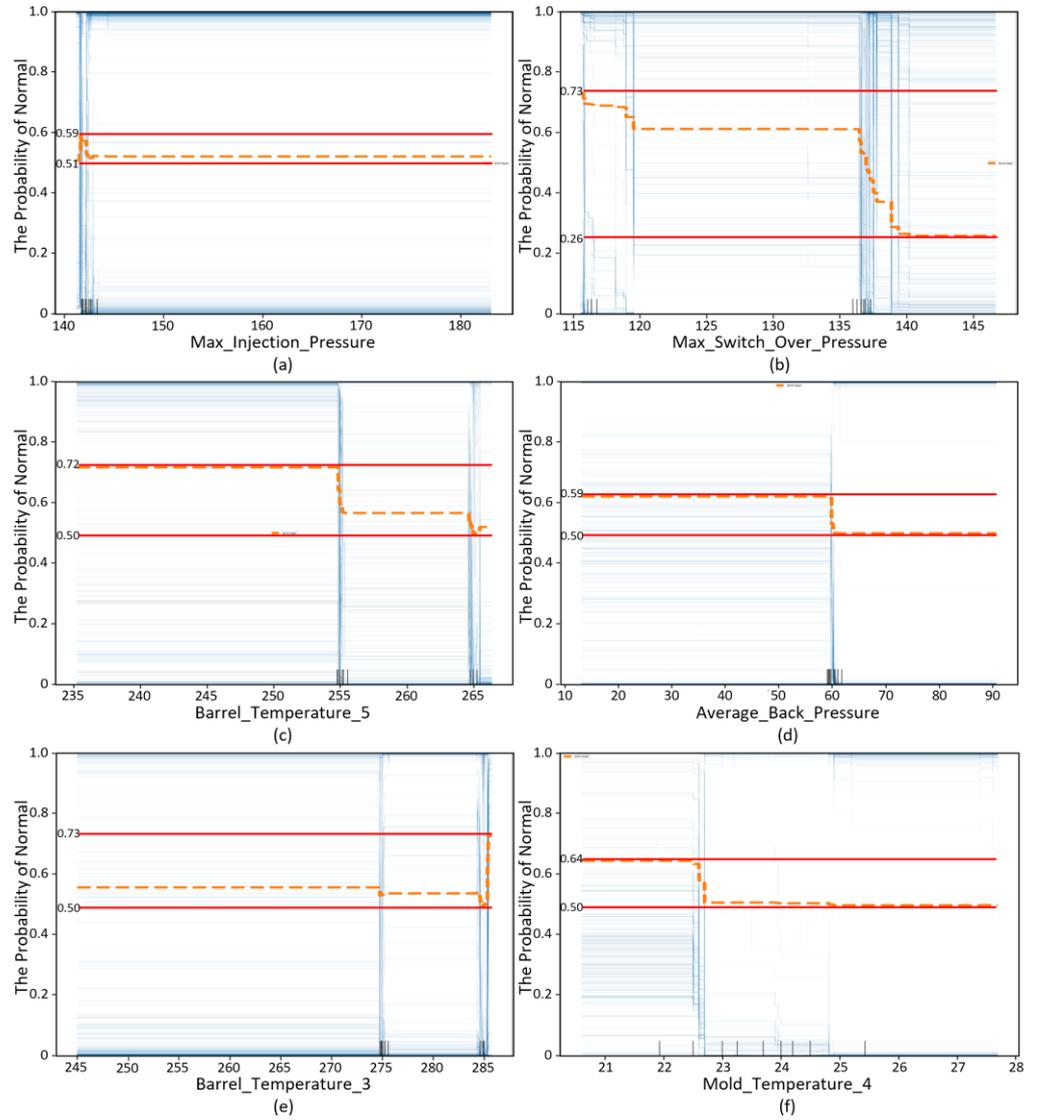

**Figure 10.** ICE Plots of LightGBM

To validate the methodology, the test dataset presented in Table 12 is utilized. The test dataset is not oversampled to reflect the low defect rate of the actual process. Subsequently, the optimal control range specified in Tables 10 and 11 is applied, and only the products produced within this range are selected. The defect rate from the test data set is compared with the original defect rate to determine whether the process has improved. The validation results are presented in Table 12.

**Table 12.** Validation Results

|               | XGBoost |        | Defect rate (%) |
|---------------|---------|--------|-----------------|
|               | Normal  | Defect |                 |
| $\alpha = 0.05$ | 969     | 2      | 0.21            |
| $\alpha = 0.1$  | 2284    | 20     | 0.88            |
| $\alpha = 0.2$  | 2284    | 20     | 0.88            |
| Original Data | 3995    | 40     | 1.00            |
|               | LightGBM |       | Defect rate (%) |
|               | Normal  | Defect |                 |
| $\alpha = 0.05$ | N/A     | N/A    | N/A             |
| $\alpha = 0.1$  | N/A     | N/A    | N/A             |
| α = 0.2 | 2314 | 3 | 0.13 |
| Original Data | 3995 | 40 | 1.00 |

When the alpha value decreases, the defect rate also decreases because of the tight control range of the process features. In the case of LightGBM, for alpha values of 0.05 and 0.1, the defect rate cannot be calculated because no data exist in this range. This also indicates that defective products are not produced. For all six experiments, the defect rate was lower than the original defect rate of 1.00%. Based on the validation, LightGBM is better for controlling the injection molding process than XGBoost. However, both algorithms requires less than a minute to process the data.

## 5. Conclusion

This paper proposes an optimal injection molding process control model to minimize the defect rate during the injection molding process. The methodology proposed in this study selects the main features of the injection molding process and presents the control range of the main features by using XAI. To predict whether the products are defective, tree-based classifier models (XGBoost and LightGBM) are used. The main features affecting the product defectivity are selected using the SHAP algorithm. The control range of the selected main features is presented by using ICE algorithm.

A test dataset was used to verify the defect rate reduction for validation. The original dataset consisted 3,995 of normal data values and 40 defect data values. The defect rate in the original dataset was 1.00%. Using XGBoost, the improved dataset comprised 969 normal data values and 2 defect data values. The defect rate in the improved dataset was 0.21%. Using LightGBM, the improved dataset consisted of 2,314 normal data values and three defect data values. The defect rate of the improved dataset was 0.13%. The defect rates were 0.79% and 0.87%, respectively.

This study proposes an optimal model for improving product yield using injection molding process data. Compared with traditional AI approaches, XAI allows injection domain experts who may lack expertise in AI to understand the results of the methodology. As the injection molding process is not performed automatically in this study, it could help support injection engineers in improving the yield rate by providing the main features with control ranges. The study authors collaborated with LG Electronics to decrease the defect rate in the injection molding process.

This study focuses on the controllable variables in the injection molding process. The field experts from LG Electronics identified the 16 features, and excluded 29 features including time and position features. Therefore, the significance of this study lies in its ability to improve process yield by adjusting the values of the main features identified in the methodology. Also, it enables field experts to more easily understand AI predictions by providing evidence for model learning by using XAI.

Through the collaborating research projects with industries, the methodology presented in this paper is extended to the practice level. Also, process datasets other than injection molding process datasets should be conducted to expand the model to various manufacturing areas. In addition, the application of neural-network-based classification models or reinforcement learning techniques should be analyzed for automated manufacturing processes.

## Abbreviations

The following abbreviations are used in this manuscript:

| | |
|---|---|
| SHAP | Shapley Additive exPlanations |
| ICE | Individual Conditional Expectation |
| PDP | Partial Dependence Plot |
| XAI | eXplainable Artificial Intelligence |